\documentclass[conference]{IEEEtran}
\IEEEoverridecommandlockouts
\usepackage{cite}
\usepackage{amsmath,amssymb,amsfonts}
\usepackage{algorithmic}
\usepackage{graphicx}
\usepackage{textcomp}
\usepackage{xcolor}
\usepackage{comment} %added by me
\usepackage{color,soul} % TODO REMOVE

\def\BibTeX{{\rm B\kern-.05em{\sc i\kern-.025em b}\kern-.08em
    T\kern-.1667em\lower.7ex\hbox{E}\kern-.125emX}}
\begin{document}

\title{Progressive Spatio-Temporal Bilinear Network with Monte Carlo Dropout for Landmark-based Facial Expression Recognition with Uncertainty Estimation}

%Progressive Spatio-Temporal Bilinear Network with Monte Carlo Dropout for Landmark-based Facial Expression Recognition with Uncertainty Estimation
\author{\IEEEauthorblockN{Negar Heidari and Alexandros Iosifidis}
\IEEEauthorblockA{\textit{Department of Electrical and Computer Engineering, Aarhus University, Denmark} \\
\{negar.heidari,ai\}@ece.au.dk}
}

\maketitle

\begin{abstract}
Deep neural networks have been widely used for feature learning in facial expression recognition systems. However, small datasets and large intra-class variability can lead to overfitting. In this paper, we propose a method which learns an optimized compact network topology for real-time facial expression recognition utilizing localized facial landmark features.
Our method employs a spatio-temporal bilinear layer as backbone to capture the motion of facial landmarks during the execution of a facial expression effectively.
Besides, it takes advantage of Monte Carlo Dropout to capture the model's uncertainty which is of great importance to analyze and treat uncertain cases. The performance of our method is evaluated on three widely used datasets and it is comparable to that of video-based state-of-the-art methods while it has much less complexity. 
\end{abstract}

%\begin{IEEEkeywords}
%Facial expression recognition, Spatio-Temporal Bilinear Networks, Graph Convolutional Networks 
%\end{IEEEkeywords}

\section{Introduction}
Facial expression recognition (FER) has been widely studied in the past several years and it is of great importance in different areas of computer vision such as social robotics and human-computer interaction (HCI). %Various machine learning have been proposed for facial expression recognition (FER) to encode the six basic facial expressions, which are anger, sadness, happiness, fear, disgust, surprise, and the neutral expression. 
Although deep learning models have a high ability in feature learning, there are different challenges for employing them in facial expression recognition. The intra-class variability, including variations in age, gender, pose, illumination, face scale and appearance, necessitates the use of complex deep learning models to extract the most useful features for expression recognition \cite{valstar2012meta}. However, existing publicly available datasets are not large and diverse enough to train high-performing deep learning models. Thus, designing compact neural network architectures for real-time facial expression recognition that can achieve high performance is of great importance. %Additionally, considering the most informative features representing the facial expressions in each image makes a step forward towards having a more efficient FER system. 

It has been shown that the FER performance can be improved by using localized facial landmarks \cite{MollahosseiniCM16}. 
The motion of facial landmarks during the execution of a facial expression effectively represents the dynamic motion of the most informative facial parts, such as eyes, nose and mouth, for facial expressions and it is also invariant to illumination conditions and face appearance. 
However, while it has been shown that multi-modal data fusion based on facial landmarks and images or videos can improve performance of image or video-based FER \cite{JungLYPK15,Kollias915,HassaniM17a}, deep learning models employing only facial landmark features have been rarely studied. 

FER methods can be categorized in \textit{static methods}, which use an image as input to classify the facial expression depicted in it, and \textit{dynamic methods} which use videos or a sequence of images as input to classify the facial expression by considering both spatial and temporal features for classification. In this work, we focus on dynamic facial expression recognition. 
Existing deep learning approaches for dynamic FER which utilize facial landmarks, typically concatenate their coordinates over multiple frames to form a sequence of vectors to be used by Recurrent Neural Networks (RNNs) \cite{JungLYPK15}, or reorganize them to form a grid map so that they can be in a form suitable to become the input of Convolutional Neural Networks (CNNs) \cite{yan2018multi}. Therefore, these methods are not capable to capture the dynamic spatial and temporal features encoded in the facial landmarks in a sequence of frames. 
Similar to human body skeletons which are used for human action recognition (HAR) \cite{yan2018spatial, shi2019two}, facial landmarks are also non-Euclidean structured data that can be modeled by a graph in which the landmark points are the graph nodes and the relationships between them are the edges connecting graph nodes. 
Therefore, the Spatio-Temporal Graph Convolutional Networks (ST-GCNs) \cite{yan2018spatial, shi2019two}, the Progressive ST-GCN (PST-GCN) \cite{heidari2021progressive} which tries to find an optimized ST-GCN architecture, or the recently introduced Spatio-Temporal Bilinear Network (ST-BLN) \cite{heidari2020spatial} can be employed to extract informative features from a sequence of graphs, encoding facial landmarks through different time steps, for facial expression recognition.

One aspect of a real FER system that is often neglected by FER methods is that of classification uncertainty. In a real-world scenario, the FER system will analyze the facial expressions of a person and take actions which can take the form of, for example, recommendations to perform an activity. Spurious misrecognized expressions caused either by misclassification due to a low-performing model, or by false identification of an expression (e.g. sadness instead of neutral) due to high uncertainty, would lead to frustration to the user. Thus, for a FER system to be practical, it needs to be based on a high-performing model which can run in real-time and estimate the uncertainty of its predictions.

In this paper, we propose the Progressive Spatio-Temporal Bilinear Network (PST-BLN) method for facial expression recognition. PST-BLN inherits the advantage of ST-BLN \cite{heidari2020spatial} to learn graph structures at each layer of the network topology without the requirement of a pre-defined graph structure, allowing for more flexible model design. Moreover, the PST-BLN method automatically defines an optimized, compact and data-dependant network topology without the need of thorough experimentation using user-designed topologies. 
Moreover, we propose to capture the model's uncertainty by training our PST-BLN model using Monte Carlo Dropout \cite{gal2021uncertainty} for helping the users of FER system to treat uncertain cases explicitly. 

\begin{comment}
Therefore, the main contributions of this paper can be summarized as follows: 
\begin{itemize}
    \item We propose PST-BLN method to find a compact network topology for real time facial expression recognition. 
    \item We propose using Monte Carlo dropout to capture the model's uncertainty. 
\end{itemize}
The remainder of the paper is organized as follows. 
Section \ref{sec:related_works}, introduces GCNs and landmark-based FER methods as the background of our method and Section \ref{sec:proposed} describes the proposed method. The experimental results are provided in section \ref{sec:experiments}, and the conclusions are drawn in section \ref{sec:conclusion}.
\end{comment}

\section{Related work} \label{sec:related_works}

Facial landmarks have been widely used in FER methods in conjunction with other data modalities to enhance performance. Recently, many real-time facial landmark detection methods have been developed which achieve good performance in addition to their high efficiency \cite{dong2018style}.
%\cite{kartynnik2019real,xiong2013supervised,ren2014face,asthana2014incremental,dong2018style}. 
\begin{comment}
Fig. \ref{fig:LM-FER} shows a sequence of $3$ facial images of a person performing the expression ``surprise'' at three different time steps (i.e. the start (left), the apex of the expression (right) and an intermediate point in time (middle)) and the aligned landmark points for each image, which are extracted using the method in \cite{kazemi2014one}. 
\begin{figure}[!t]
    \centering
    \includegraphics[width=0.3\linewidth]{Figures/LM_4.png}
    \includegraphics[width=0.3\linewidth]{Figures/LM_6.png}
    \includegraphics[width=0.3\linewidth]{Figures/LM_14.png}
    \caption{Illustration of facial images in $3$ different time steps and their corresponding extracted landmark points which are denoted as graph nodes in a spatio-temporal graph. The images are from CK+ dataset \cite{lucey2010extended}.
    }
    \label{fig:LM-FER}
\end{figure}
\end{comment}

Recently, a GCN-based method has been proposed in \cite{ngoc2020facial} which uses only facial landmarks for facial expression recognition. 
In \cite{ngoc2020facial}, the landmark extractor \cite{dong2018style} was adopted to extract accurate $2$D coordinates of $68$ landmark points from each facial image in an image sequence. The extracted landmarks were modeled by a directed spatio-temporal graph which is constructed using landmark points as nodes and triangle meshes among all landmarks, built by Delaunay method, as edges. 
Inspired by methods recently proposed for skeleton-based human action recognition, like the DGNN \cite{shi2019skeleton}, the FER method \cite{ngoc2020facial} also employs a multi-layer spatio-temporal GCN model to extract features from the spatio-temporal facial landmark graph and introduces the extracted features to a fully connected classification layer to predict the facial expression. 

\section{Proposed method}\label{sec:proposed}
This section describes the proposed PST-BLN method for dynamic landmark-based facial expression recognition. The description starts with the graph construction procedure, followed by the description of the Spatio-Temporal Bilinear Layer (ST-BLL) and the proposed PST-BLN method. The combination of PST-BLN with Monte Carlo Dropout for estimating the model's uncertainty is finally described.  

\subsection{Spatio-temporal graph construction} 
By extracting the facial landmarks of all the images in a sequence, a spatio-temporal graph $\mathcal{G} = (\mathcal{V} ,\mathcal{E})$ can be constructed where $\mathcal{V}$ is the node set of $2$D coordinates of the facial landmarks and $\mathcal{E}$ is the set of graph edges encoding spatial and temporal connections between the landmarks through different time steps. 
In this work, we adopted the Dlib's facial landmark extractor \cite{kazemi2014one} to extract accurate $2$D coordinates of $68$ landmark points from each facial image.
It has been shown in \cite{ngoc2020facial} that the landmarks of the outer region of the face do not contain informative features for different facial expressions. 
%and they might have negative impact on the model's performance. 
Therefore, we remove the first $17$ facial landmarks of each image and keep only the 51 landmarks carrying features of the key facial parts for facial expression recognition. 
Facial landmarks in each graph are normalized by subtracting the central landmark (nose). 
The triangle meshes among all landmarks obtained by Delaunay method make the spatial graph edges. The central node (nose) is set as the master node which is connected to all other graph nodes. The temporal graph edges connect each landmark into its corresponding landmark in its previous and subsequent frames.

We utilize the edge features of the graph which encode the motion of the facial muscles instead of the landmark coordinates. Each graph edge is bounded by two graph nodes and it can be defined as a feature vector representing both the length and direction information. As an example, we define the feature vector of a graph edge with source node $v_i = (x_i,y_i) $ and target node $v_j = (x_j, y_j)$ as $e_{ij}=(x_i - x_j, y_i - y_j) $. Therefore, each image in the sequence is modeled by a graph with $E$ spatial edges and the PST-BLN receives as input a tensor $\mathbf{X} \in \mathbb{R}^{F \times T \times E}$ encoding a sequence of $T$ spatial graphs expressing the connections of the graph edges. $F$ denotes the feature dimension of each edge feature $e_{ij}$.

\subsection{Spatio-Temporal Bilinear Layer}
The ST-BLL is composed of a bilinear transformation and a temporal convolution. 
The bilinear transformation receives as input the representations for the $E^{(l-1)}$ facial graph edges at layer $l-1$, denoted by $\mathbf{H}^{(l-1)}$, and transforms them by using a learnable weight matrix $\mathbf{W}^{(l)}$ as follows:
\begin{equation}
    \mathbf{H}_{s}^{(l)} = ReLU\left( \mathbf{U}^{(l)}\mathbf{H}^{(l-1)}\mathbf{W}^{(l)} \right),
    \label{eq:bl_Convs}
\end{equation}
where $\mathbf{U}^{(l)} \in \mathbb{R}^{E^{(l)} \times E^{(l-1)}}$ is a learnable matrix indicating the spatial weighted connections between the facial graph edges. 
This matrix is initialized randomly and it is optimized in an end-to-end manner jointly with the parameters of the entire network. 
Unlike GCN layers which use the graph Adjacency matrix in the spatial graph convolution, ST-BLL allows for freely deciding the dimensions of matrix $\mathbf{U}$. This means that ST-BLL allows for aggregating (or expanding) information of the graph edges leading to $E^{(l)} < E^{(l-1)}$ (or $E^{(l)} > E^{(l-1)}$, respectively). In this paper, we chose to keep the number of graph edges constant for all ST-BLLs, and for notation simplicity we use $E$ hereafter.

\begin{figure}[]
    \centering
    \includegraphics[width=0.9\linewidth]{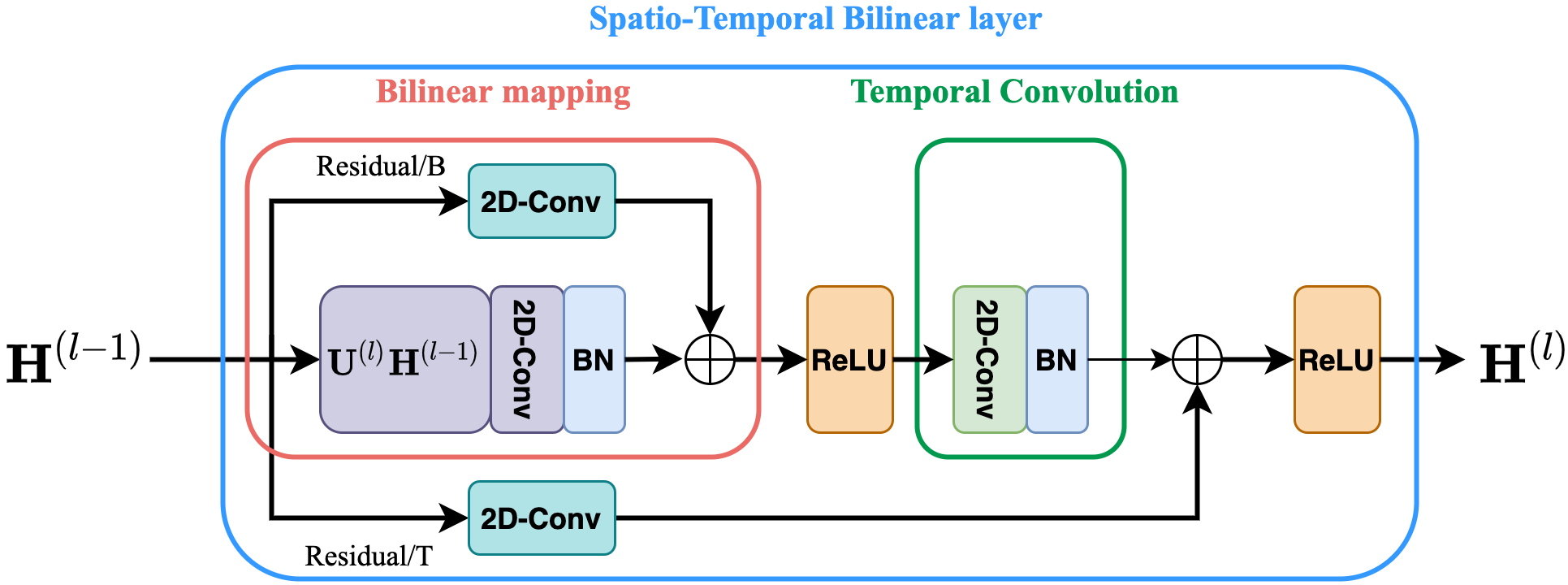}
    \caption{Illustration of spatio-temporal bilinear layer $l$. It receives $\mathbf{H}^{(l-1)}$ of size $F^{(l-1)} \times T^{(l-1)} \times E$ as input and applies bilinear projection and temporal convolution to produce the output representation $\mathbf{H}^{(l)}$ of size $F^{(l)} \times T^{(l)} \times E$.
    %Bilinear mapping and temporal convolution both are standard 2D convolutions depicted as 2D-Conv blocks. The bilinear layer utilizes $F^{(l)}$ convolutions with filters of size  $F^{(l-1)} \times 1 \times 1 $ to transform the data and the temporal convolution keeps the number of channels unchanged and aggregates the temporal features in $K$ frames by applying a convolution with $F^{(l)}$ filters of size $F^{(l)} \times K \times 1 $. 
    The bilinear mapping block and the temporal convolution block are both followed by batch-normalization (BN) and ReLU activation function. 
    %The residual connections Residual/B and Residual/T add the layer's input to the output of bilinear mapping and temporal convolution, respectively.
    }
    \label{fig:ST-BLL}
\end{figure}

The spatially transformed feature tensor $\mathbf{H}_s^{(l)} \in \mathbb{R}^{F^{(l)} \times T^{(l)} \times E}$ with $F^{(l)}$ feature dimensions is introduced to the temporal convolution, which captures the motion of the facial muscles taking place in each facial expression by propagating the edge features of each spatial graph through the time domain using a standard 2D convolution with a predefined kernel size $K \times 1$ aggregating edge features in $K$ consecutive frames. 
The structure of the ST-BLL is shown in Fig. \ref{fig:ST-BLL}. Each layer of the network is euqipped by two residual connections to stabilize the model by adding the input to the output of the bilinear mapping and the temporal convolution. The temporal convolution block is followed by batch normalization and ReLU activation function.

\subsection{Progressive spatio-temporal bilinear network (PST-BLN)}
%Inspired by PST-GCN method \cite{heidari2021progressive} defining a data-driven method for learning an optimized and problem-dependant topology for ST-GCN, we propose PST-BLN method to build a compact ST-BLN topology in terms of both width and depth progressively.
%PST-BLN receives the tensor $\mathbf{X} \in \mathbb{R}^{F \times T \times E}$ as input and builds a ST-BLN network from scratch. 
%\hl{Didn't go through the following two paragraphs. We need to describe the multiplication with $W^{(l)}$ as a projection and not as a convolution.} 

A ST-BLN model is composed of several ST-BLLs for feature extraction and one fully connected layer for classification. A network with $l$ ST-BLLs, employs the global average pooling after the $l^{th}$ layer to produce a feature vector of size $F^{(l)} \times 1$. The feature vector is introduced to a fully connected layer which maps features from $F^{(l)}$ to $C$ dimensional subspace to classify features into $C$ different classes. 

Let us assume that a ST-BLN with $l-1$ layers has been already built, and the method proceeds in building the $l^{th}$ layer. 
In practice, the bilinear projection and temporal convolution in \ref{eq:bl_Convs} are standard 2D convolutions with filters of sizes $F^{(l)} \times 1 \times 1$, and $F^{(l)} \times F^{(l)} \times K \times 1$, respectively. $F^{(l)}$ denotes the number of output channels in the $l^{th}$ layer and $K$ denotes the kernel size in the temporal convolution. 
The residual connections are also standard 2D convolutions which transform the input data of the layer with filters of size $F^{(l)} \times 1 \times 1$ to have the same dimension as the layer's output. 
When the method starts building the $l^{th}$ layer, the number of output channels in all the 2D convolutions is set to a predefined fixed number $F^{(l)}=b$ and at each iteration, it is increased by $F^{(l)}= F^{(l)} + b$.
While all the model's parameters in the previously built layers are initialized by the finetuned weights, the newly added neurons to the network are initialized randomly and all the model parameters are fine-tuned in an end-to-end manner using back-propagation.
The layer's width progression at iteration $t$ is evaluated according to the model's performance in terms of categorical loss value on training data, i.e. $\alpha_{w}= (\mathcal{L}_{t-1}^{(l)}- \mathcal{L}_{t}^{(l)}) / \mathcal{L}_{t-1}^{(l)}$. 
$\mathcal{L}_{t-1}^{(l)}$ and $\mathcal{L}_{t}^{(l)}$ denote the model's loss value at iterations $t-1$ and $t$, respectively. 
If $\alpha_{w} < \epsilon_{w}$ with $\epsilon_{w} > 0$, it shows that increasing the layer's width doesn't improve the model's performance anymore and the method stops progression in that layer.
Otherwise, the newly added parameters are saved and the next iteration starts increasing the layer's width by adding $b$ more output channels to the filters of all the 2D convolutions in that layer. 

This process repeats iteratively until the performance converges in that layer.
After building each layer of the network, the method evaluates the model's depth progression using the rate of improvement in model's performance, i.e. $\alpha_{d} = (\mathcal{L}^{(l-1)}- \mathcal{L}^{(l)}) / \mathcal{L}^{(l-1)}$, in terms of categorical loss value on training data. $\mathcal{L}^{(l-1)}$ and $\mathcal{L}^{(l)}$ denote the model's loss value before and after adding the new layer to the network, respectively. 
When $\alpha_{d}< \epsilon_{d}$ with $\epsilon_{d} > 0$, the method stops depth progression and the newly added layer is removed. Finally, all the model's parameters are fine-tuned together and the method returns the optimized topology for the ST-BLN model and its performance on training and validation data. 

%\vspace{-1cm}
\subsection{PST-BLN with Monte Carlo Dropout to model uncertainty}
People of different ages, genders and cultural backgrounds have different levels of expressiveness, and they perform or interpret the facial expressions in different ways. 
Although the output of a classification model (softmax scores) encodes the predictive (pseudo-)probabilities of the model, it has been shown that even models with high softmax outputs can be uncertain about their predictions \cite{gal2016dropout}. 
Since facial expression datasets are small in size, regularization of the network parameters is needed to prevent overfitting. To address this, we add a dropout layer after each ST-BLL built by the proposed method using a dropout rate $p$ of $0.2$. This choice also allows us to use Monte Carlo Dropout \cite{gal2016dropout} to capture the uncertainty of the model during inference. This is very helpful for the users of the FER system to interpret the facial expression of a sample when the model is uncertain about its prediction. 
The main idea of Monte Carlo Dropout is to use dropout not only in the training phase, but also during the inference. 
Since dropout randomly switches off a subset of neurons in each layer, it can be interpreted as a Bayesian approximation of the Gaussian process. Every time the model provides classification result with activated dropout layers, its outputs are obtained from slightly different models with different sets of activated neurons and each of these models can be treated as a Monte Carlo sample. By repeating the inference for an input facial spatio-temporal graph with an activated dropout, the outputs of the PST-BLN are combined as an ensemble of different PST-BLN models and the variance in the outputs are used to capture the classification uncertainty. 
%The only drawback of this method is that the inference time is related to the number of inference repeats and as we get more Monte Carlo samples, we have more inference time. 

\section{Experimental results}\label{sec:experiments}
\subsection{Datasets}
The preformance of our method has been evaluated on the following three widely used datasets: 
\\
\textbf{CK+}\cite{lucey2010extended, kanade2000comprehensive}: The Extended Cohn–Kanade (CK+) contains $327$ videos of $7$ different emotional classes, starting from a neutral expression to peak expression. 
Similar to most methods using this dataset for evaluation, we select the first frame and the last three frames (including the peak expression) of each sequence for landmark extraction. Besides, the subjects are divided into $10$ groups for 10-fold cross-validation. 
\\
\textbf{Oulu-CASIA} \cite{zhao2011facial}: The Oulu-CASIA dataset consists of $2,880$ image sequences of $80$ subjects, captured under three different illumination conditions and using two different imaging systems; near-infrared (NIR) and visible light (VIS). We used the $480$ image sequences captured by the VIS system under normal indoor illumination and we divided the subjects into $10$ groups for 10-fold cross validation. In each image sequence, we used the last three frames, including the peak expression, and the first frame showing the neutral expression.  
\\
\textbf{AFEW}\cite{dhall2012collecting, dhall2014emotion}: 
The Acted Facial Expressions in the Wild (AFEW) dataset is a more challenging dataset for landmark extraction methods compared to the CK+ and Oulu-CASIA. It consists a set of video clips collected from movies with actively moving faces in different illumination and environmental conditions. In some frames of each video where head pose is not frontal, the landmark extraction methods confront challenges to detect the face and extract its landmarks. Therefore, only a subset of video frames which provide meaningful facial landmark features are used. The dataset is divided into three sets, train, validation and test, with labels for the test set not being publicly available. Therefore, models are trained on the training set and evaluated on the validation set. 

\subsection{Experimental setup}
The experiments are conducted with GeForce RTX 2080 GPUs, SGD optimizer with weight decay of $0.0005$ and momentum of $0.9$ and cross entropy loss function. The models are trained on AFEW dataset for $300$ epochs on $4$ GPUs with learning rate of $0.01$ and batch size of $64$. For CK+ and Oulu-CASIA datasets, the models are trained for $400$ epochs with learning rate of $0.1$ and batch size of $128$. The PST-BLN method is trained with block sizes of $5$ and layer/block thresholds of $0.0001$ for all three datasets. 

Since AFEW dataset is challenging and it does not have sufficient amount of data for training the model, we adopted data augmentation to expand the dataset size by $14$ times. 
First, for each video we extracted the landmarks from $150$ frames which are sampled at same time intervals and when the number of frames with meaningful landmarks are less than $150$, we repeat the frames by tiling method. 
After landmark extraction, similar to \cite{JungLYPK15, ngoc2020facial}, we added three different Gaussian noises to facial landmarks, then we applied random rotation to the noised data followed by random flipping to each sequence.

\subsection{Performance evaluation}
The performance of the proposed method is compared with both video-based and landmark-based state-of-the-art methods on AFEW, Oulu-CASIA and CK+ datasets in Tables \ref{table:AFEW-ACC}, \ref{table:Oulu-CASIA-ACC}, \ref{table:CK+-ACC}, respectively. 
In each table, the state-of-the-art methods are divided into two groups. The first group contains the CNN-based or RNN-based methods which use videos or image sequences as the main data stream for training the model while some of these methods such as \cite{jung2015joint, zhang2017facial} also utilize the landmark data in conjunction with video/image sequence to highlight the most important parts of the facial images and improve the performance. The second group contains the GCN-based methods which only use facial landmarks.
To the best of our knowledge, the only GCN-based method that has been proposed for facial expression recognition is DGNN \cite{ngoc2020facial} which is an extension of \cite{shi2019skeleton} method for facial expression recognition. To show the effectiveness of our proposed model compared to other GCN-based networks, we also include in the comparisons the well-known GCN-based methods such as ST-GCN \cite{yan2018spatial} and AGCN \cite{shi2019two} to evaluate their performance on the landmark-based facial expression recognition task. Video-based methods train CNN and RNN-based architectures such as VGG16, LSTM, C3D, and they have the best performance on these datasets. However, the number of parameters of some of these methods is not reported in the corresponding papers. 

The ST-BLN model is composed of 7 ST-BLLs with output dimensions of $\{8, 16, 16, 32, 32, 64, 64\}$, respectively and a fully connected layer for classification. This model topology is the same as DGNN's topology, and in order to have a fair comparison, we modified the topology of ST-GCN and AGCN models to have the same number of layers and layer dimensions as DGNN and ST-BLN models. 
While ST-GCN and AGCN methods utilize the landmark coordinates, or graph node features, and the squared Adjacency matrix of the graph in the spatial convolution, ST-BLN and PST-BLN utilize only the edge features of the graph.
DGNN utilizes both node features and edge features encoded by a directed graph. 

\begin{table}
\caption{Comparison of video/image-based and landmark-based methods on the validation set of AFEW dataset}\label{table:AFEW-ACC}
\begin{center}
\vspace*{-5mm}
\resizebox{0.9\linewidth}{!}{
    \begin{tabular}{l|c|c|c}
        \hline
        Method  & Acc(\%) & \#Params  & Data type \\ %
		\hline
		SSE-HoloNet \cite{hu2017learning}  & 46.48 & - & Video \\ 
		VGG-LSTM  \cite{vielzeuf2017temporal}  & 48.60 & -  & Video \\ %VGG16-LSTM
		C3D-LSTM  \cite{vielzeuf2017temporal}  & 43.20 & - & Video\\ %C3D-LSTM
		C3D-GRU  \cite{lee2019visual}  & 49.87 & - & Video\\ 
        \hline
		ST-GCN \cite{yan2018spatial} & 28.17 & 131.3k& Landmark\\ 
		AGCN \cite{shi2019two} & 24.21 & 143.7k & Landmark\\ 
		DGNN \cite{shi2019skeleton} & 32.64 & 538k & Landmark\\ 
		\hline \hline
		\bf{ST-BLN w/MCD} & \bf{36.11} & \bf{132.3k} & Landmark\\ %
		\bf{ST-BLN wo/MCD} & 34.13 & 132.3k & Landmark\\ %
		\bf{PST-BLN w/MCD} & \bf{33.33} & \bf{10.8k} & Landmark\\ %
		\bf{PST-BLN wo/MCD} & 30.15 & 10.8k & Landmark\\ %
		\hline
    \end{tabular}}
\end{center}
\vspace*{-5mm}
\end{table}
\begin{figure}[]
    \centering
    \includegraphics[width=0.65\linewidth]{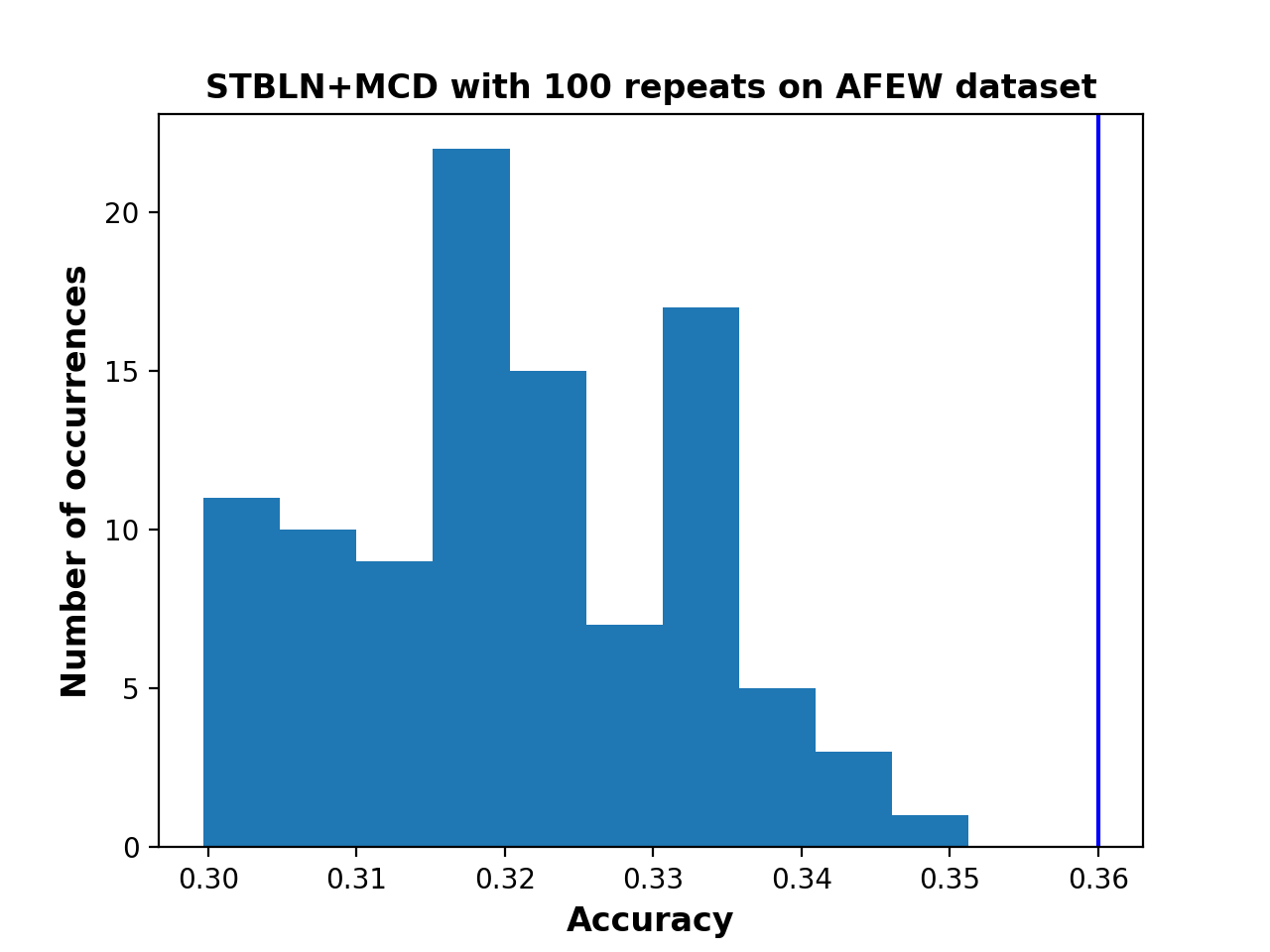}
    \caption{The distribution of 100 classification accuraceis obtained by the ST-BLN w/MCD method on AFEW dataset. The vertical line in the left side indicates the classification accuracy obtained by the ensembled predictions.}
    \label{fig:AFEW_MCDO_ACCdist}
\end{figure}

\begin{figure}
    \centering
    \includegraphics[width=0.3\linewidth]{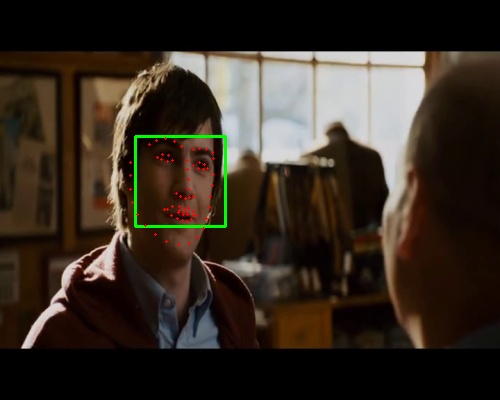}
    \includegraphics[width=0.3\linewidth]{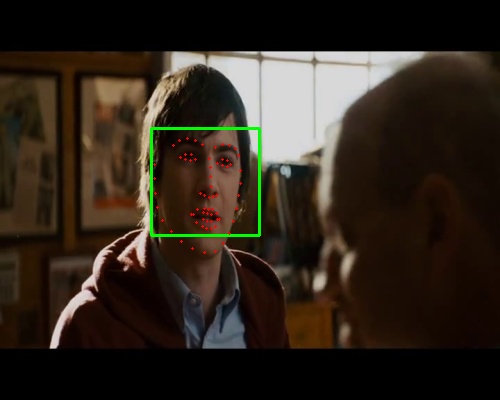}
    \includegraphics[width=0.3\linewidth]{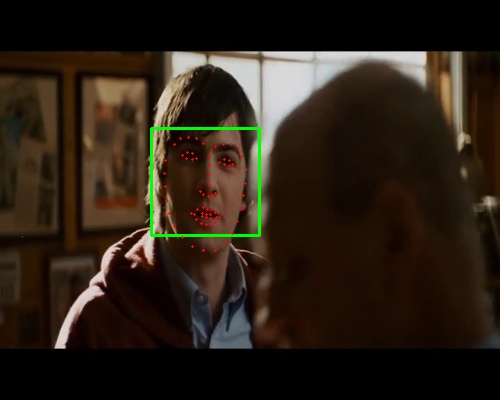}
    \includegraphics[width=0.9\linewidth]{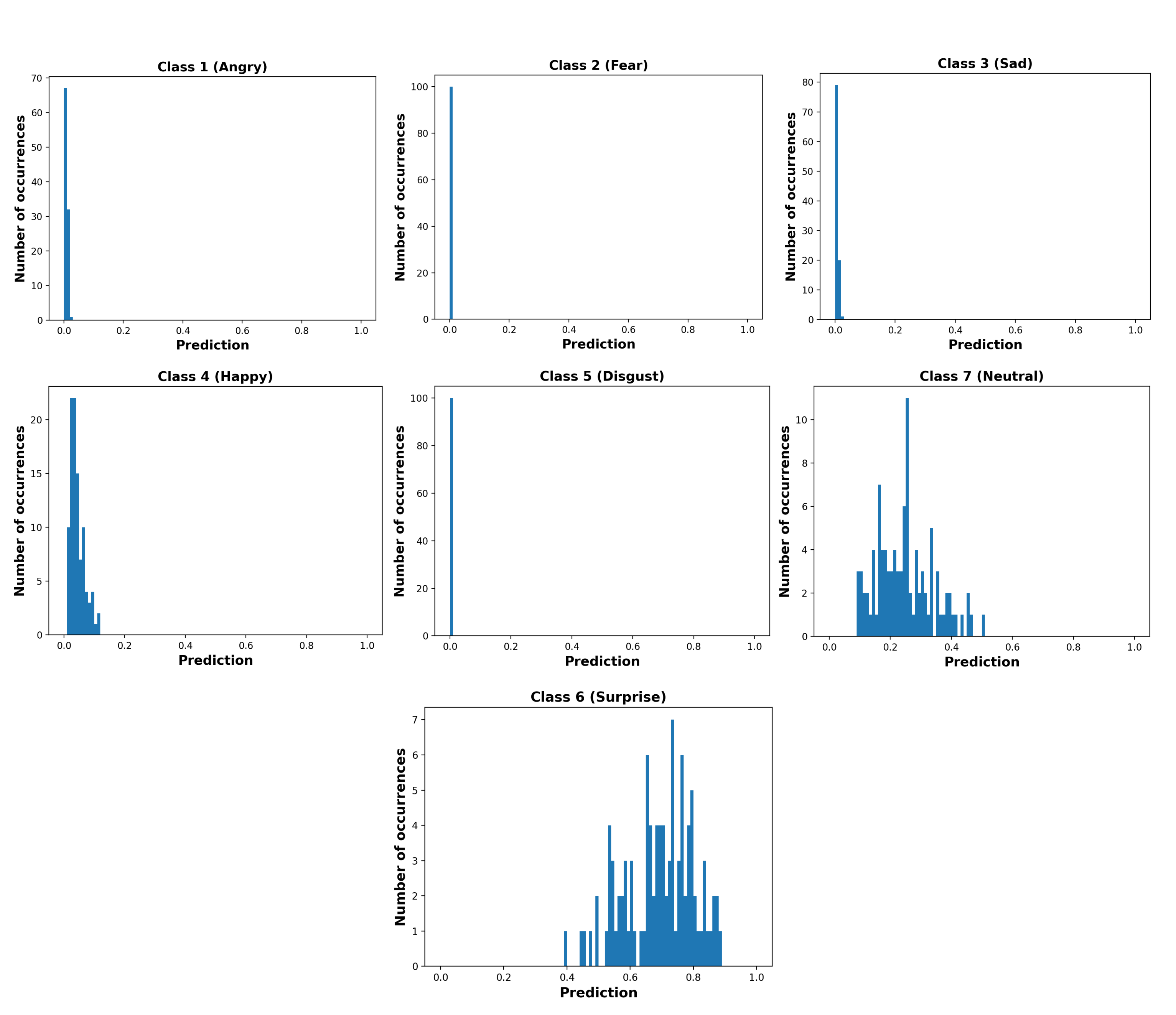}
    \caption{Illustration of $3$ frames of a sample video in AFEW dataset expressing `Surprise`, top row, and the distribution of $100$ predictions for each class, obtained by our proposed ST-BLN model.}
    \label{fig:MCDO_Surprise_dist}
    \vspace*{-5mm}
\end{figure}

Experimental results on AFEW dataset indicate that ST-BLN outperforms all the baseline GCN-based methods, ST-GCN and AGCN, with a large margin while they have quite similar model complexity in terms of number of parameters. Compared to DGNN, ST-BLN has improved the classification performance by $4\%$ while it has $4$ times less number of parameters. 
PST-BLN which is trained with block sizes of $5$ and layer/block thresholds of $0.0001$, found an optimized topology for this dataset which is composed of $6$ ST-BLLs with output sizes $\{15, 10, 15, 5, 5, 10\}$, respectively. This optimized model outperforms DGNN, ST-GCN and AGCN models with only $10.8 k$ parameters which are $49$ times less than those of DGNN. 

To capture the model's uncertainty, we evaluated both ST-BLN and PST-BLN models with activated dropout layers during the inference and we repeated the inference for $100$ times on each sample to get $100$ different prediction vectors. ST-BLN w/MCD and PST-BLN w/MCD denote the model's classification accuracy obtained as the mean of $100$ different predictions and ST-BLN wo/MCD and PST-BLN wo/MCD report the classification accuracy obtained by performing the inference only once.
The results show that the model achieves better performance when it ensembles the predictions of $100$ models rather than performing the inference only once. 
To calculate the model's uncertainty on a dataset, we calculate the classification accuracy over $100$ runs. 
Fig. \ref{fig:AFEW_MCDO_ACCdist}, shows the distribution of $100$ classification accuracy values of the ST-BLN w/MCD on AFEW dataset. The mean and standard deviation of this distribution are $32.09$, $1.18$, respectively. The classification accuracy obtained by the ensembled predictions is shown by a vertical line in the left side of the figure which is $36.11\%$ and it is around $4\%$ better than the mean accuracy. 

Additionally, our proposed method gives the user the possibility of visualizing the model's uncertainty on an individual sample base. As an example, we evaluated the ST-BLN model on a video sample of class Surprise from the AFEW dataset. Fig. \ref{fig:MCDO_Surprise_dist} illustrates $3$ frames of this video with their extracted facial landmarks and also the prediction distribution for each expression class. This figure shows that the model classifies this sample correctly in the Surprise class with mean probability of $0.69$ while it is uncertain about it. Considering the sample frames in the top row of the figure, it can be seen that it is a hard example to classify and based on the prediction distributions, this example can also be classified in Neutral and Happy classes with mean probabilities of $0.24$, $0.04$, respectively.
The variance of the model predictions of each class can be interpreted as the model's uncertainty on that class. Therefore, the model's uncertainty on classes Surprise, Neutral and Happy is $0.1$, $0.9$, $0.02$, respectively.

\begin{table}
\caption{Comparison of video-based and landmark-based methods on Oulu-CASIA dataset using 10-fold cross validation}\label{table:Oulu-CASIA-ACC}
\begin{center}
\vspace*{-5mm}
\resizebox{0.9\linewidth}{!}{
    \begin{tabular}{l|c|c|c}
        \hline
        Method  & Acc(\%) & \#Params  & Data type \\ %
		\hline
		DTAN \cite{jung2015joint} & 74.38 & - & Video\\ %IntraFace
		DTGN \cite{jung2015joint} & 74.17 & 177.6k & Landmark\\%IntraFace
		DTAGN \cite{jung2015joint} & 81.46 & -  & Video + Landmark\\ %IntraFace \cite{xiong2013supervised} --> landmark extractor which is also used for face detection
        PPDN \cite{zhao2016peak} & 84.59 & 6.8m & Video  \\
        PHRNN-MSCNN \cite{zhang2017facial}  & 86.25 & 1.6m & Video + Landmark \\ %IntraFace \cite{xiong2013supervised} --> landmark extractor 
        DCPN \cite{yu2018deeper}  & 86.23 & - & Video \\ %MTCNN \cite{zhang2016joint} --> face detection and alignment
        CDLM \cite{kuo2018compact} & 91.67 & 2.7m & Video \\ %IntraFace --> used for face detection
        \hline
        ST-GCN \cite{yan2018spatial} & $77.08$ & 131.3k & Landmark \\ 
        AGCN \cite{shi2019two} & $75.62$ & 143.7k  & Landmark \\ 
        DGNN \cite{shi2019skeleton} & $81.46$ & 535,7k & Landmark \\ 
        \hline \hline
        \bf{ST-BLN w/MCD} & $\bf{83.54}$ & \bf{132.3k}  & Landmark \\ %
        \bf{ST-BLN wo/MCD} & $82.08$ & 132.3k  & Landmark \\ %
        \bf{PST-BLN w/MCD} & $\bf{79.79}$ & $\bf{7.59k}$ & Landmark \\ %
        \bf{PST-BLN wo/MCD} & $78.74$ & $7.59$k & Landmark \\ %
        \hline
    \end{tabular}}
\end{center}
\vspace*{-5mm}
\end{table}

\begin{table}
\caption{Comparison of the video-based and landmark-based methods on CK+ dataset using 10-fold cross validation}\label{table:CK+-ACC}
\begin{center}
\vspace*{-5mm}
\resizebox{0.9\linewidth}{!}{
    \begin{tabular}{l|c|c|c}
        \hline
        Method  & Acc(\%) & \#Params  & Data type\\ %
		\hline
		DTAN \cite{jung2015joint} & 91.44 & - & Video \\
		DTGN \cite{jung2015joint} & 92.35 & 177.6k & Landmark\\
		DTAGN \cite{jung2015joint} & 97.25 & -  & Video + Landmark\\ 
        PPDN \cite{zhao2016peak} & 99.3 & 6.8m & Video \\
        PHRNN-MSCNN \cite{zhang2017facial} & 98.5 & 1.6m & Video + Landmark \\ 
        DCPN \cite{yu2018deeper} & 99.6 & - & Video\\ 
        CDLM \cite{kuo2018compact} & 98.47 & 2.7m & Video\\
        \hline
        ST-GCN \cite{yan2018spatial} & 93.64  & 131.3k & Landmark\\ 
        AGCN \cite{shi2019two} & 94.18  & 143.7k & Landmark\\ 
        DGNN \cite{shi2019skeleton} & 96.02 & 535,7k & Landmark \\ 
        \hline \hline
        \bf{ST-BLN w/MCD} & $\bf{95.47} $ & \bf{132.3k} & Landmark \\ %
        \bf{ST-BLN wo/MCD} & $93.19$ & 132.3k & Landmark\\ %
        \bf{PST-BLN w/MCD} & $\bf{93.34}$ & \bf{9.79k} & Landmark \\ %
        \bf{PST-BLN wo/MCD} & $93.1$ & 9.79k& Landmark \\ %
        \hline
    \end{tabular}}
\end{center}
\vspace*{-5mm}
\end{table}

The mean classification performance of the models over all folds is reported for Oulu-CASIA and CK+ datasets. Since the PST-BLN method finds a different model topology for each fold of the data, we report the average number of parameters of 10 optimized models. 
Experimental results on Oulu-CASIA dataset show that the proposed ST-BLN model outperforms all the landmark-based methods while it has $4$ times less number of parameters compared to the DGNN method. PST-BLN is trained separately for each of the $10$ folds of the data and the average number of parameters is reported which is around $70$ times less than DGNN and $17$ times less than ST-BLN. 
Although the PST-BLN does not outperform the state-of-the-art methods, it is competitive while being much more compact.

ST-BLN outperforms the ST-GCN and AGCN on CK+ dataset while it has competitive performance compared to DGNN with around $4$ times less number of parameters. The optimized topology PST-BLN achieves similar performance to the ST-GCN and AGCN with around $13$ and $14$ times less number of parameters. 
It should be noted that the reported number of parameters in all the tables corresponds only to the neural network models. As we use \cite{kazemi2014one} for landmark detection, other methods such as \cite{jung2015joint, zhang2017facial, kuo2018compact} also utilize IntraFace \cite{xiong2013supervised} landmark extractor and \cite{yu2018deeper} employs MTCNN \cite{zhang2016joint} for face detection and alignment as a pre-processing step. 
The results of ST-BLN w/MCD and PST-BLN w/MCD on Oulu-CASIA and CK+ datasets also confirm that repeating the inference with activated dropouts and ensembling the results, improves the classification performance.

\section{Conclusion}\label{sec:conclusion}
In this paper, we proposed a method which builds an optimized and compact spatio-temporal bilinear network topology for facial expression recognition by employing the localized facial landmarks instead of videos or image sequences. While our method has achieved comparable performance to more complex state-of-the-art methods, it captures the model's uncertainty using Monte Carlo Dropout technique which allows the user to analyze the model's prediction for different cases and take desired action.

\section*{Acknowledgment}
This work was supported by the European Union’s Horizon 2020 Research and Innovation Action Program under Grant 871449 (OpenDR). This publication reflects the authors’ views only. The European Commission is not responsible for any use that may be made of the information it contains.

\bibliographystyle{IEEEbib}
\bibliography{bibliography}

\end{document}